\documentclass[10pt,twocolumn,letterpaper]{article}

\usepackage{iccv}
\usepackage{times}
\usepackage{epsfig}
\usepackage{graphicx}   
\usepackage{color}
\usepackage{pifont}
\usepackage{amsmath}
\usepackage{amssymb}
\usepackage{balance}
\usepackage{eso-pic}
\usepackage{multirow}
\usepackage{array}
\usepackage{caption}
\usepackage[pagebackref=true,breaklinks=true,letterpaper=true,colorlinks,bookmarks=false]{hyperref}

\iccvfinalcopy 

\def\iccvPaperID{3402} 

\ificcvfinal\fi
\begin{document}

\title{Occlusion-shared and Feature-separated Network\\ for Occlusion Relationship Reasoning}



\author{Rui Lu$^{1}$\hspace{0.5cm}Feng Xue$^{1}$\hspace{0.5cm}Menghan Zhou$^{1,3}$\hspace{0.5cm}Anlong Ming$^{1}$\hspace{0.5cm}Yu Zhou$^{2,*}$\\
{$^{1}$Beijing University of Posts and Telecommunications, Beijing, China}\\
{$^{2}$Huazhong University of Science and Technology, Wuhan, China}\\
{$^{3}$Lenovo Research, Beijing, China}\\
{\tt\small{\{lurui,xuefeng,mal\}@bupt.edu.cn} zhoumh3@lenovo.com yuzhou@hust.edu.cn}
}

\maketitle
\renewcommand{\thefootnote}{\fnsymbol{footnote}}
\footnotetext[1]{Corresponding Author}

\pagestyle{empty}  
\thispagestyle{empty} 

\begin{abstract}
Occlusion relationship reasoning demands closed contour to express the object, and orientation of each contour pixel to describe the order relationship between objects.
Current CNN-based methods neglect two critical issues of the task:
(1) simultaneous existence of the relevance and distinction for the two elements, i.e, occlusion edge and occlusion orientation; and
(2) inadequate exploration to the orientation features.
For the reasons above, we propose the Occlusion-shared and Feature-separated Network (OFNet).
On one hand, considering the relevance between edge and orientation,
two sub-networks are designed to share
the occlusion cue.
On the other hand, the whole network is split into two paths to learn the high-level semantic features separately.
Moreover, a contextual feature for orientation prediction is extracted, which represents the bilateral cue of the foreground and background areas.
The bilateral cue is then fused with 
the occlusion cue
to precisely locate the object regions.
Finally, a stripe convolution is designed to further aggregate features from surrounding scenes of the occlusion edge.
The proposed OFNet remarkably advances the state-of-the-art approaches on PIOD and BSDS ownership dataset. The source code
is available at \url{https://github.com/buptlr/OFNet}.

\end{abstract}

\section{Introduction}
\label{sec:intro}
Reasoning the occlusion relationship of objects from monocular image is 
fundamental in computer vision and mobile robot applications, such as \cite{Jacobson2012An,Ayvaci2011Detachable,Sargin2009Probabilistic,Marshall1996Occlusion,Stein2009Occlusion}. 
Furthermore, 
it can be regarded as crucial elements for scene understanding and visual perception\cite{Yu2016Similarity,Yu2016Human,Zhou2014ONLINE,NIPS2012_4792,Ming2016Monocular}, such as object detection, image segmentation and 3D reconstruction
\cite{Gao2011A,Alper2012Detachable,Zhang2015Monocular,He2017Mask,Shen2015DeepContour,Xue2019A}. 
From the perspective of the observer, occlusion
relationship reflects relative depth difference between objects in the scene. 

\begin{figure}[t]
\centering
\includegraphics[width=1\linewidth]{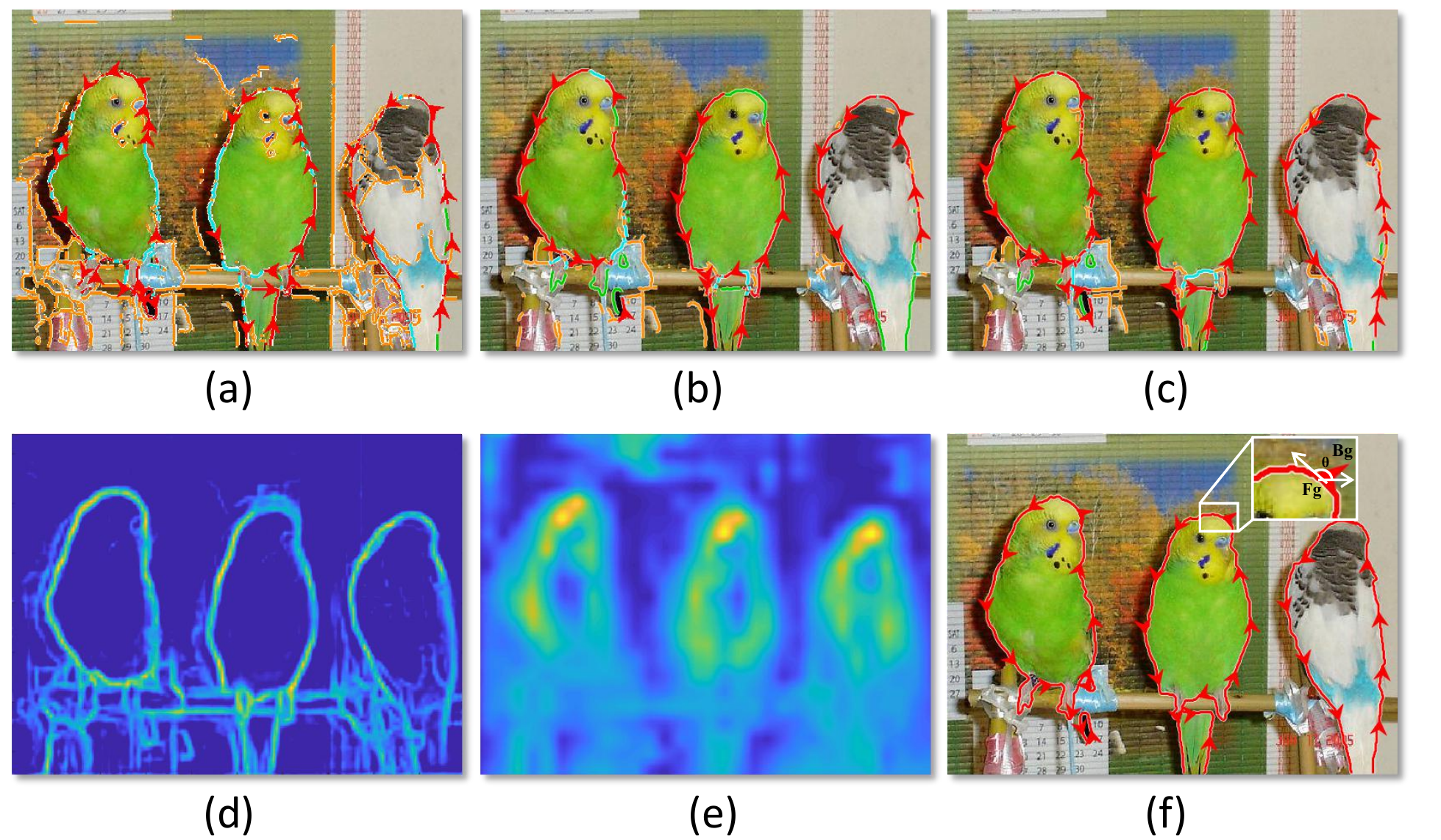}
\caption{
(a) visualization result of DOC-HED, (b) visualization result of DOOBNet, (c) visualization result of ours, (d) the occlusion cue, (e) the bilateral feature, (f) visualization result of ground truth. 
Occlusion relationship (the red arrows) is represented by orientation $\theta\in(-\pi,\pi]$ (tangent direction of the edge), using the "left" rule where the left side of the arrow means foreground area.
Notably, "red" pixels with arrows: correctly labeled occlusion boundaries; "cyan": correctly labeled boundaries but mislabeled occlusion; "green": false negative boundaries; "orange": false positive boundaries (Best viewed in color). 
}
\label{Fig:demo}
\end{figure}

\begin{figure*}[t]
\centering
\includegraphics[width=0.9\linewidth]{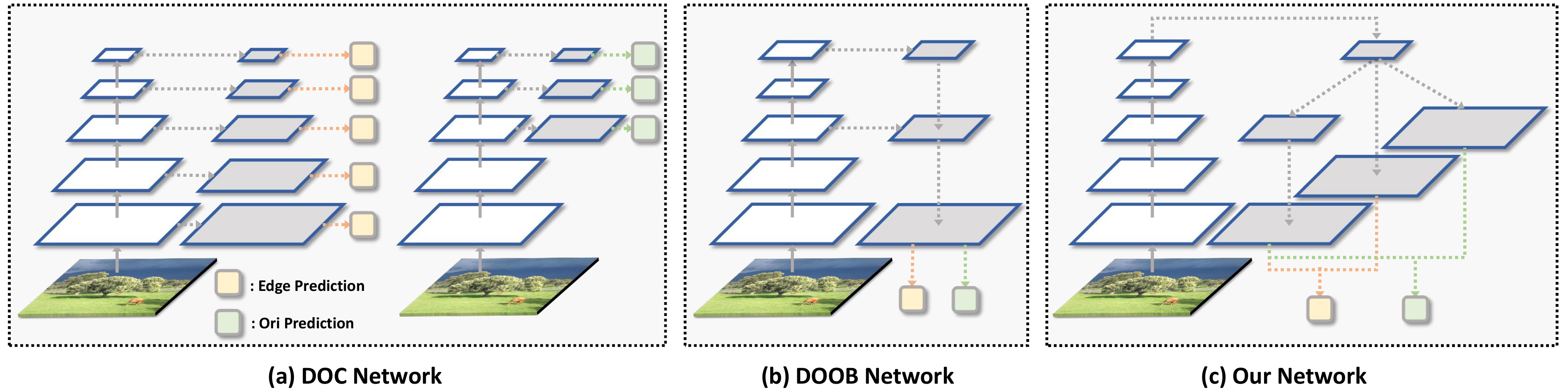}
\caption{
The schematic demonstration of the high-level feature propagation process of the state-of-the-arts and ours. (a) indicates the two separate stream networks employing side-outputs of various layers. (b) presents the single stream network sharing decoder features. (c) shows our network which captures contextual features for specific tasks and shares decoder features. 
}
\label{Fig:netdemo}
\end{figure*}

Previously,
a number of influential studies infer the occlusion relationship by designing hand-crafted features,
e.g. \cite{Hoiem2007Recovering,Ren2006Figure,Saxena2005Learning,Ma2017Object,inproceedings,Zhou2018Learning}.
Recently, driven by Convolutional Neural Networks (CNN),
several deep learning based 
approaches outperform traditional methods at a large margin.
DOC \cite{Peng2016DOC} specifies a new representation for occlusion relationship,
which decomposes the task into the occlusion edge classification and the occlusion orientation regression.
And it utilizes two networks for these two sub-tasks, respectively.
DOOBNet \cite{Wang2018DOOBNet} employs an encoder-decoder structure to obtain multi-scale and multi-level features. It shares backbone features with two sub-networks and simultaneously acquires 
the predictions.

In occlusion relationship reasoning,
\emph{
the closed
contour is employed to express the object,
and the orientation values of the contour pixels are employed to describe the order relationship between the foreground and background objects.}
We observe that two critical issues have rarely been discussed.
Firstly, the two elements, i.e, occlusion edge and occlusion orientation,
have the relevance and distinction simultaneously.
They both need the occlusion cue,
which describes the location of the occluded 
background, as shown in Fig.\ref{Fig:demo} (d).
Secondly, the high-level features for orientation prediction are not fully revealed.
It needs additional cues from foreground and background areas (shown in Fig.\ref{Fig:demo} (e)).
Consequently, existing methods are limited in reasoning accuracy.
Compared with our approach (shown in Fig.\ref{Fig:demo} (c)), previous works \cite{Peng2016DOC,Wang2018DOOBNet} (shown in Fig.\ref{Fig:demo} (a)(b)) exist false positive and false negative detection of edge, 
as well as false positive prediction of orientation.

Aiming to address the two issues above, 
and boost 
the occlusion relationship reasoning,
a novel Occlusion-shared and Feature-separated Network (\textbf{OFNet}) is proposed.
As shown in Fig.\ref{Fig:netdemo} (c),
considering the relevance and distinction between edge and orientation, our network is
different from the other works (shown in Fig.\ref{Fig:netdemo} (a)(b)).
Two separate network paths share
the occlusion cue
and encode different high-level features.
Furthermore, a contextual feature for orientation prediction is extracted,
which is called the bilateral feature.
To learn the bilateral feature,
a Multi-rate Context Learner (\textbf{MCL}) is proposed.
The learner has different scales of receptive field so that it can 
fully sense the two objects, i.e, 
the foreground and background objects,
fundamentally assisting the occlusion relationship reasoning.
To extract the feature more accurately,
the Bilateral Response Fusion (\textbf{BRF}) is proposed to fuse 
the occlusion cue
with the bilateral feature from \textbf{MCL},
which can precisely locate the areas of foreground and background.
To effectively infer the occlusion relationship by the special orientation features,
a stripe convolution is designed to replace the traditional plain convolution,
which elaborately integrates the bilateral feature to distinguish the foreground and background areas.
Experiments prove that we achieve SOTA performance on both PIOD \cite{Peng2016DOC} and BSDS ownership \cite{Ren2006Figure} dataset.
The main contributions of our approach lie in:
\begin{itemize}
\item 
The relevance and distinction between occlusion edge and occlusion orientation are re-interpreted.
The two sub-tasks share the occlusion cues,
but separate the contextual features.
    
\item
The bilateral feature is proposed, 
and two particular modules are designed to obtain
the specific features, 
i.e, Multi-rate Context Learner (\textbf{MCL}) and Bilateral Response Fusion (\textbf{BRF}).

\item
To elaborately infer the occlusion relationship,
a stripe convolution is designed to further aggregate the feature
from surrounding scenes of the contour.
\end{itemize}


\section{Related Work}
\label{sec:related}


{\bf{Contextual Learning}} plays an important role in scene understanding and perception \cite{DBLP:journals/corr/ChenPSA17,Wang2018Understanding}. 
At first, Mostajabi et al. \cite{Mostajabi2015Feedforward} utilize multi-level, zoom-out features to promote feedforward semantic labeling of superpixels.
Meanwhile, Liu et al. \cite{Liu2015ParseNet} propose a 
simple FCN architecture to add the global 
context for semantic segmentation.
Afterwards, Chen et al. \cite{Chen2018DeepLab} apply the  \emph{Atrous Spatial Pyramid Pooling} to extract dense features and encode image context at multi-scale.




{\bf{Multi-level Features}} are extracted from different layers, which are widely used in image detection \cite{Long2014Fully,Ronneberger2015U,Wei2016Object,Wang2018DeepFlux}. Peng et al. \cite{Peng2017Large} fuse feature maps from multi-layer with refined details. Shrivastava et al. \cite{DBLP:journals/corr/ShrivastavaSMG16} adopt lateral connections to leverage top-down context and bottom-up details. 


{\bf{Occlusion Relationship Representation}} has evolved overtime from triple points and junctions representation \cite{Hoiem2011Recovering,Ren2006Figure} to pixel-based representation \cite{Teo2015Fast,Peng2016DOC}.
The latest representation \cite{Peng2016DOC}
applies a binary edge classifier to determine whether the pixel belongs to an occluded edge, and a continuous-valued orientation variable is proposed
to indicate the occlusion relationship by the left-hand rule \cite{Peng2016DOC}.

\begin{figure}[t]
\centering
\includegraphics[width=1\linewidth]{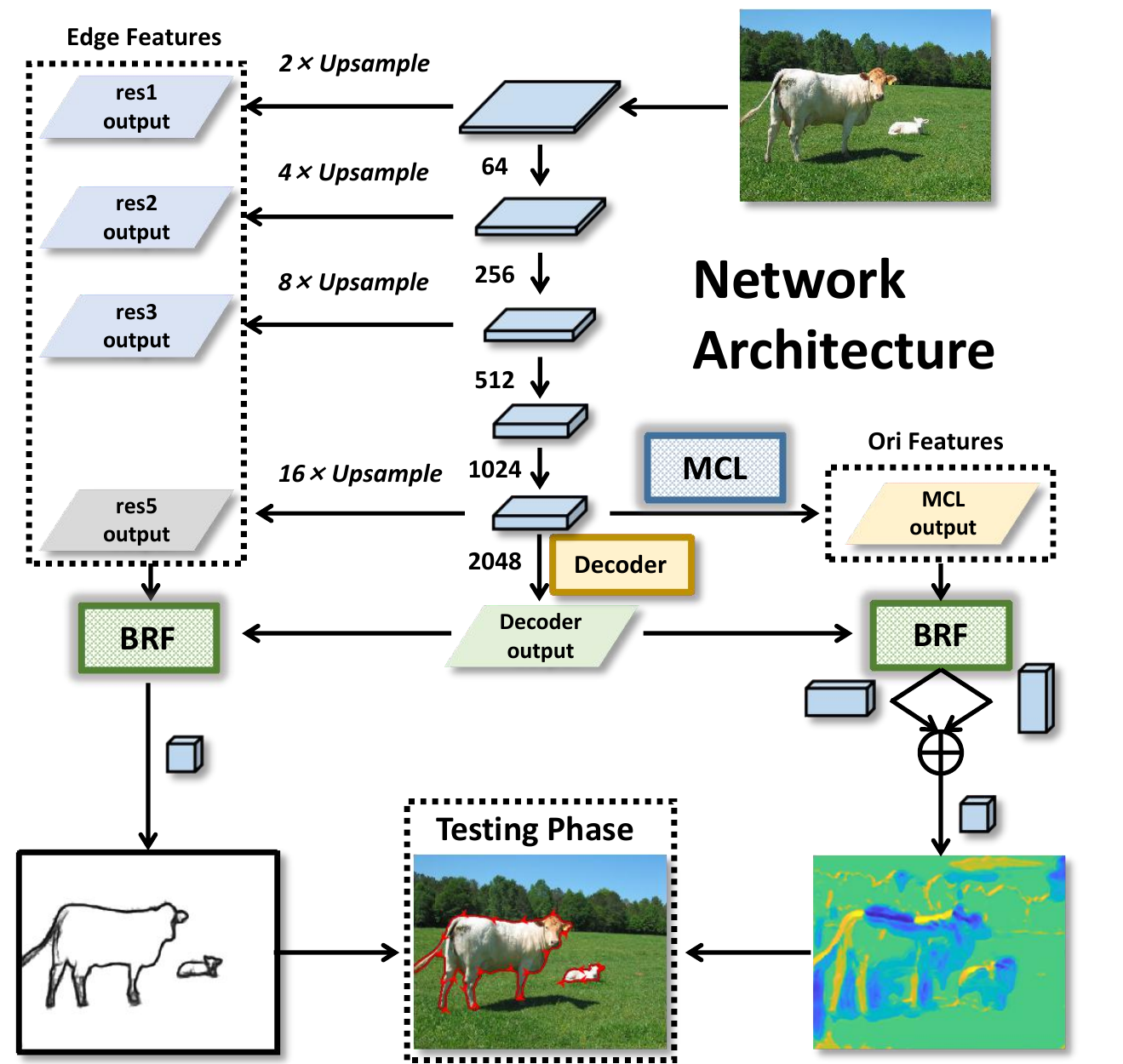}
\caption{
Illustration of our proposed network architecture. The length of the block expresses the map resolution and the thickness of the block indicates the channel number.
}
\label{Fig:network}
\end{figure}

\section{OFNet}
\label{sec:network}
Two elements of occlusion relationship reasoning, i.e, edge and orientation,
are in common of 
necessity for the occlusion cue
while differing in the utilization 
of specific contextual features.
In this section, 
a novel Occlusion-shared and Feature-separated Network (\textbf{OFNet}) is proposed. 
Fig.\ref{Fig:network} illustrates the pipeline of the proposed OFNet,
which consists of a single stream backbone and two parallel paths, i.e, edge path and orientation path.

Specifically, for the edge path (see Sec.\ref{ssec:edgepath}), 
a structure similar to \cite{Lu2019Context} is employed to extract consistent and accurate occlusion edge,
which is fundamental for occlusion reasoning.
For the orientation path (see Sec.\ref{ssec:oripath}),
to learn more sufficient cues near the boundary for occlusion reasoning,
the high-level bilateral feature is obtained,
and a Multi-rate Context Learner (\textbf{MCL}) is proposed to extract the feature (see Sec.\ref{ssec:learner}).
To enable the learner to locate the foreground and background areas precisely,
a Bilateral Response Fusion module (\textbf{BRF}) is proposed to fuse the bilateral feature and the occlusion cue (see Sec.\ref{ssec:fusion}).
Furthermore, a stripe convolution is proposed to infer the occlusion relationship elaborately
(see Sec.\ref{ssec:inference}).

\subsection{Edge Path}
\label{ssec:edgepath}
The occlusion 
edge expresses the position of objects, and defines the boundary location between the bilateral regions.
It requires reserved resolution of the original image to provide the accurate location and 
large receptive field to perceive the mutual constraint of pixels on the boundary.

We adopt the module proposed in \cite{Lu2019Context}, which has a high 
capability to capture accurate location cue and sensitive perception of the entire object.
In \cite{Lu2019Context},
the low-level cue from the first three side-outputs preserves 
the original 
size of the input image and encodes 
abundant spatial information.
Without losing resolution,
the large 
receptive field is achieved via dilated convolution \cite{Yu2015Multi} after \emph{res50} \cite{He_2016_CVPR}.
The Bilateral Response Fusion (\textbf{BRF}) shown in Fig.\ref{Fig:network} is presented to compensate the precise 
position for high-level features and suppress the clutter of non-occluded pixels for low-level 
features.
Different from \cite{Lu2019Context}, 
we employ an additional convolution block to refine the contour,
and integrate specific task features provided by diverse channels.
Besides, this well-designed convolution block
eliminates the gridding artifacts \cite{Yu2017Dilated} caused by the dilated convolution in high-level layers.

The resulting edge map embodies low-level and high-level features, which guarantees the consistency 
and accuracy of the occlusion edge. 
Specifically, the edge 
path provides complete and continuous contour, which makes up 
the object 
region. 
The object region is delineated by a set of occlusion edges.


\subsection{Orientation Path}
\label{ssec:oripath}

For the orientation 
path, we innovatively introduce the bilateral feature,
which is conducive to describe the order relationship.
Specifically, the bilateral 
feature represents information of surrounding scenes,
which includes sufficient ambient context to deduce whether it belongs to the foreground or background areas. 


\subsubsection{Multi-rate Context Learner}
\label{ssec:learner}

Bilateral feature characterizes the relationship between the foreground and background areas.
To infer the occlusion relationship between objects,
the sufficient receptive field for the objects with different sizes is essential.



\begin{figure}[t]
\centering
\includegraphics[width=0.9\linewidth]{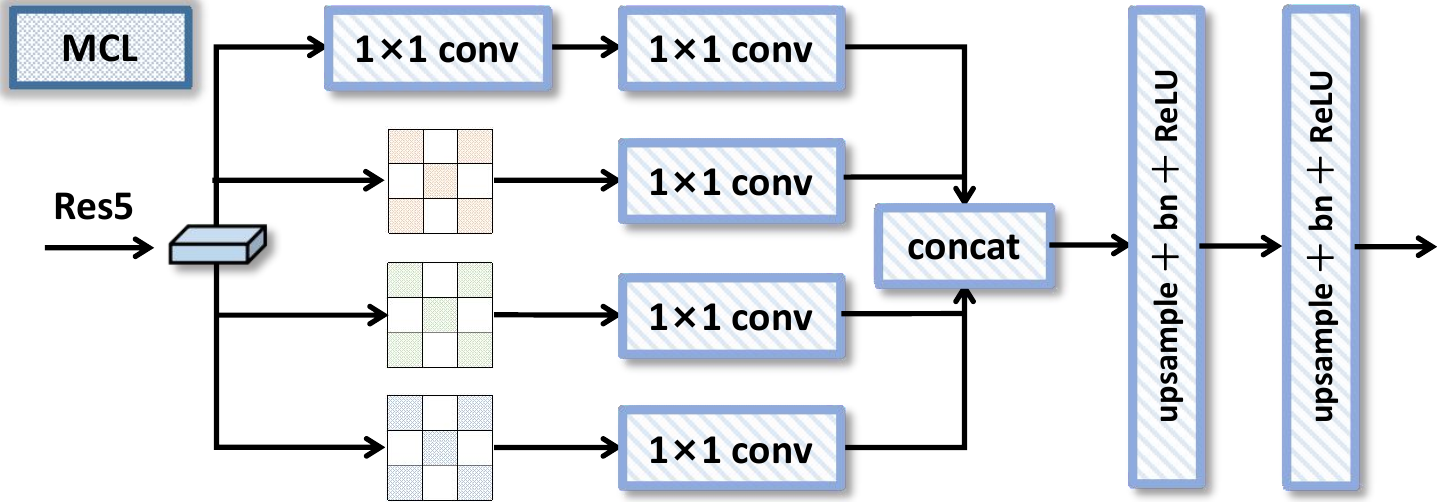}
\caption{
Illustration of our proposed Multi-rate Context Learner (\textbf{MCL}).
The \textbf{MCL} module includes 3 dilated convolutions with kernel size of \emph{3$\times$3} and dilation rate of 6, 12 and 18, respectively.
}
\label{Fig:module1}
\end{figure}

To perceive the object with various ranges and learn the bilateral feature,
the Multi-rate Context Learner (\textbf{MCL}) is designed,
which consists of three components, as shown in Fig.\ref{Fig:module1}.
Firstly, the high-level semantic cue is convolved by multiple dilated convolutions,
which allows the pixels on the edge to perceive the foreground and background objects as completely as possible.
The dilated convolutions have kernel size of \emph{3$\times$3} with various dilation rates. 
With various dilated rates of the dilated convolutions,
the learner is able to perceive the scene cue at different scales 
from the foreground and background areas,
which is beneficial to deduce which side of the region is in front.
Secondly, an element-wise convolution module, i.e., \emph{1$\times$1} conv, is used to integrate the scene cue between various channels and promote cross-channel bilateral feature aggregation at the same location.
Compared to dilated convolution,
the element-wise convolution module retains the local cues near the contour.
Besides, it greatly clarifies 
occlusion cue
and bilateral cue in occlusion reasoning.
The function $Dilated()$ represents the dilated convolution and the function $Conv_1()$ represents the \emph{1$\times$1} conv.
$X_i$ is the input of convolution.
$W_i$ is the convolution layer parameters to be learned.
Thirdly, the \emph{1$\times$1} conv is once again applied to normalize the values nearby the contour,
where the bilateral cue is further enhanced and other irrelevant cues are suppressed.
The \textbf{MCL} learns the cues of the foreground and background objects.
The feature map of bilateral cue, $i.e.,\{B\}$, is denoted as:
\begin{equation}
\label{Eq:MCL}
\{B\} = Conv_1(\sum_{i=1}^{3}Dilated(X_i,\{W_i\}) + Conv_1(X_4,\{W_4\}))
\end{equation}

\begin{figure}[t]
\centering
\includegraphics[width=0.9\linewidth]{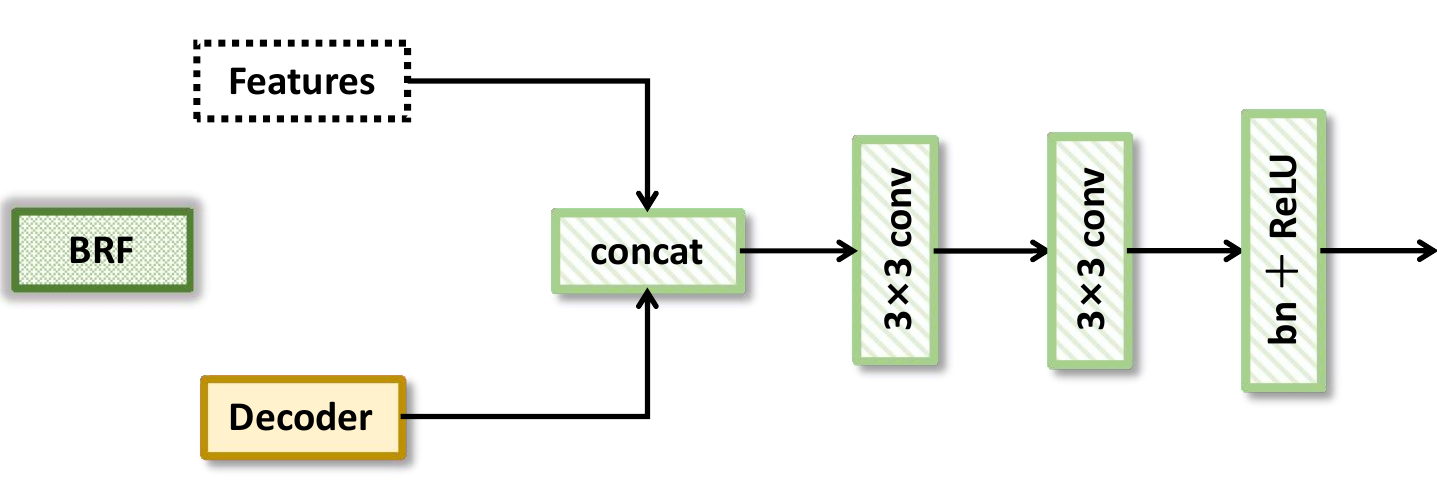}
\caption{
Illustration of our proposed Bilateral Response Fusion (\textbf{BRF}).
}
\label{Fig:module2}
    \end{figure}


{\bf{Difference with \textbf{ASPP}:}} Notably, our \textbf{MCL} module is inspired by the “Atrous Spatial Pyramid Pooling” (ASPP) \cite{Chen2018DeepLab}, 
but there exists several differences. 
Firstly, we add a parallel element-wise convolution module, 
which additionally gains local cues of the specific region. 
It compensates for the deficiency that dilated convolution is not sensitive to nearby information. 
Secondly, the convolution blocks after each branch remove the gridding artifacts \cite{Yu2017Dilated} caused by the dilated convolution. Thirdly, the \emph{1$\times$1} conv can adjust channel numbers and explore relevance between channels.



\subsubsection{Bilateral Response Fusion}
\label{ssec:fusion}

\begin{figure}[t]
\centering
\includegraphics[width=0.9\linewidth]{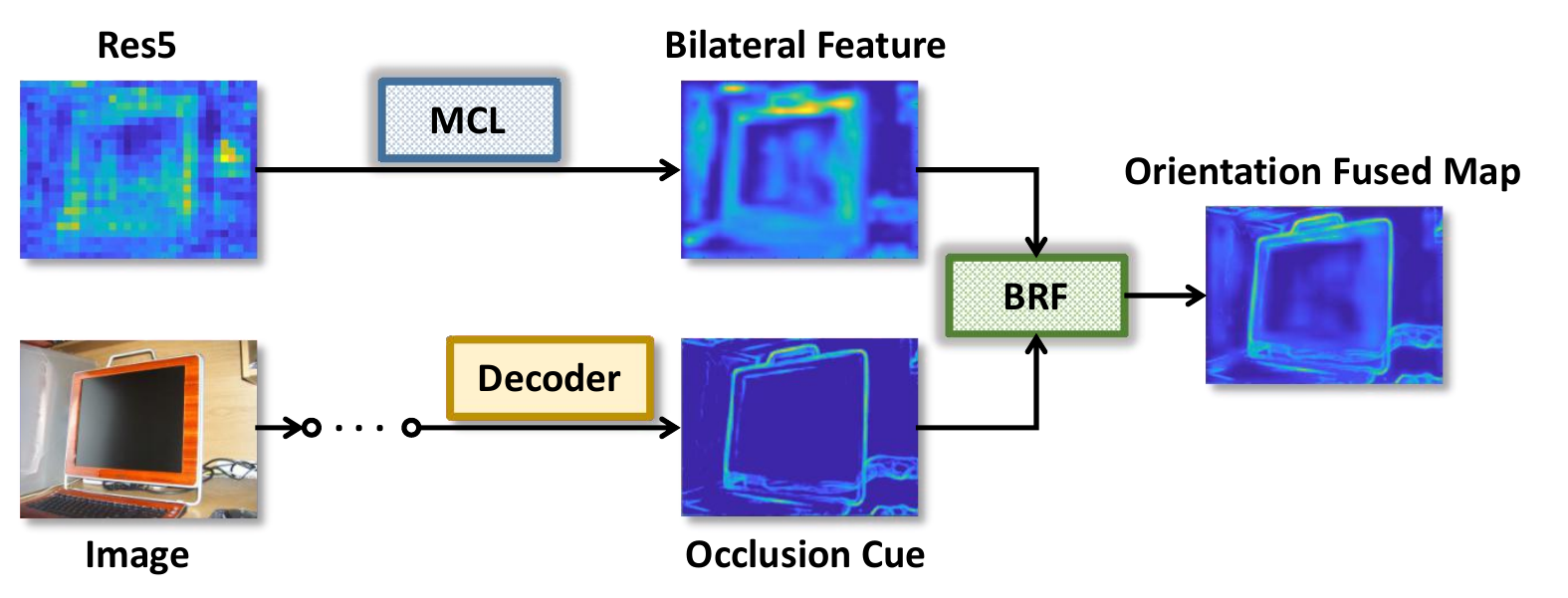}
\caption{
The test demo of the generation of orientation fused map. We acquire the fused map by adopting \textbf{BRF} to complement bilateral feature with 
occlusion
feature.
}
\label{Fig:aspp}
\end{figure}

Respectively,
the bilateral cue obtained by the method in Sec.\ref{ssec:learner} discriminates which side of the contour belongs to foreground area,
the occlusion cue
obtained through the decoder represents the location information of the boundary.
As shown in Fig.\ref{Fig:aspp},
after the bilinear upsampling,
bilateral feature
is hard to locate the exact location of the contour.
Hence, to sufficiently learn the feature for occlusion relationship reasoning,
more precise location of object region is demanded,
which is provided by
the occlusion cue
from the decoder.
Thus, it is necessary to introduce clear contour to describe the areas of the foreground and background objects,
thereby extracting the object features more accurately.

The Bilateral Response Fusion (\textbf{BRF}), shown in Fig.\ref{Fig:module2}, is proposed to fuse these two disparate streams of features,
i.e. the bilateral map $\{B\}$ and 
occlusion
map $\{D\}$.
The unified orientation fused map of ample bilateral response and emphatic 
occlusion
is formed,
which is denoted as $\{F\}$, where $Conv_3()$ represents the \emph{3$\times$3} conv:
\begin{equation}
\label{Eq:BRF}
\{F\} = Conv_3(Conv_3(Concat(\{B\}, \{D\})))
\end{equation}
$\{\emph{F}\}$ denotes the feature map generated by \textbf{BRF} module, and each element
of the set
is a feature map.
Subsequently, $\{\emph{F}\}$ has \emph{224$\times$224} spatial resolution and is taken as the input of the \textbf{Occlusion Relationship Reasoning} module (Sec.\ref{ssec:inference}), as shown in Fig.\ref{Fig:aspp}.
Through \textbf{BRF}, 
the occlusion
feature is effectively combined with the bilateral feature.
For occlusion relationship reasoning,
the fused orientation map not only possesses the boundary location between two objects with occlusion relationship,
but also own contextual information of each object.
The \textbf{BRF} module provides adequate cues for the following feature learning module to infer the foreground and background relationship.
Besides, by integrating bilateral feature, 
the scene cue near the contour is enhanced.

\subsubsection{Occlusion Relationship Reasoning}
\label{ssec:inference}

By utilizing the \textbf{MCL} and \textbf{BRF},
the bilateral feature is learned and fused,
an inference module is necessary to determine the order of the foreground and background areas,
which makes full use of this feature.
Existing method \cite{Wang2018DOOBNet} utilizes \emph{3$\times$3} conv to learn the features.
This small convolution kernel only extracts the cues at the local pixel patch,
which is not suitable to infer the occlusion relationship.
The reason is that the tiny perceptive field is unable to perceive the learned object cue.
Thus, a large convolution kernel is necessary for utilizing the bilateral feature,
which is able to perceive surrounding regions near the contour.

\begin{figure}[t]
\centering
\includegraphics[width=0.9\linewidth]{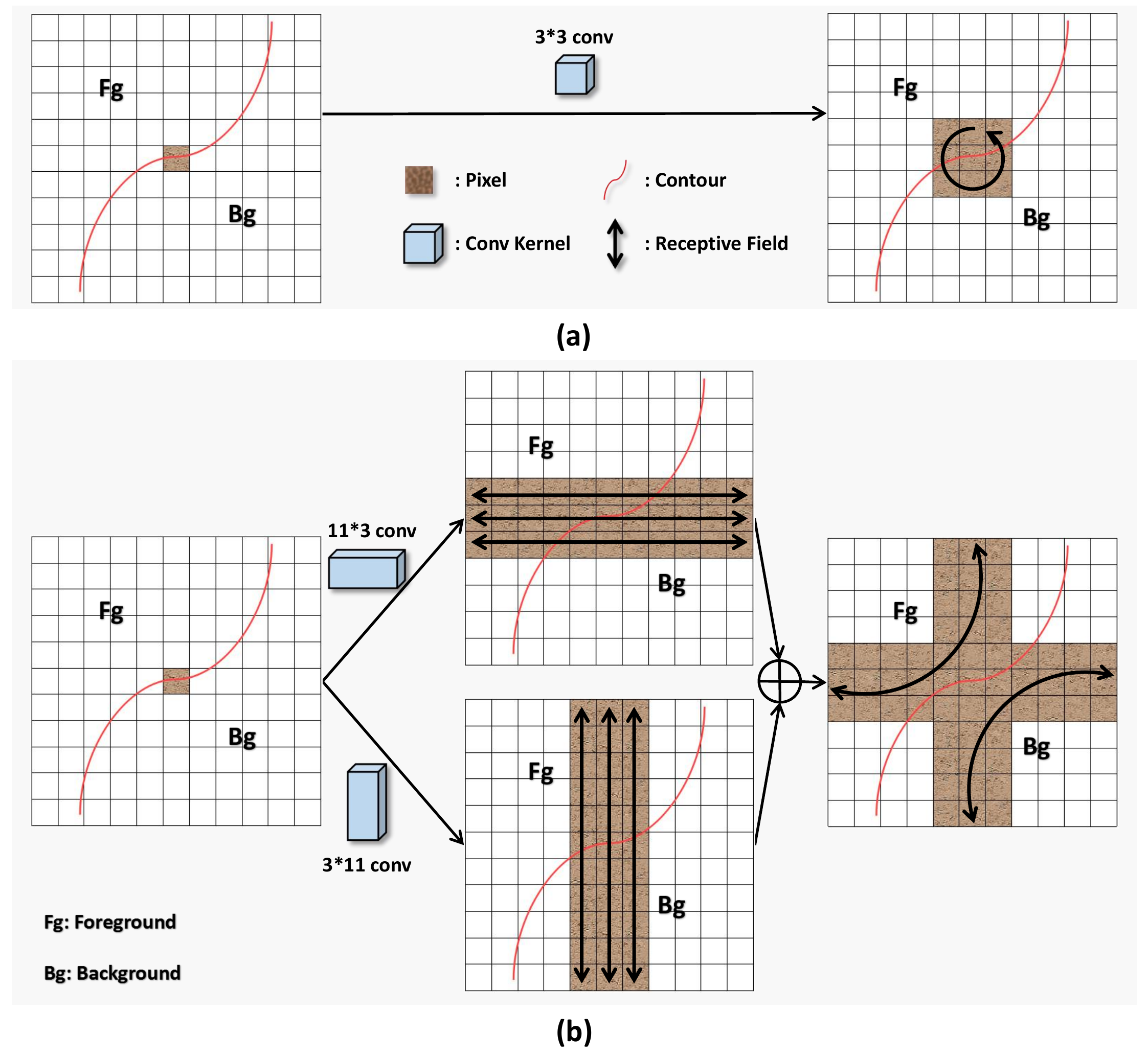}
\caption{
The schematic illustration of how orientation information propagates in the feature learning phase. (a) the plain convolution. (b) the stripe convolution.
}
\label{Fig:stripe}
\end{figure}

Nevertheless, large convolution kernels are computation demanding and memory consuming.
Instead, two stripe convolutions are proposed,
which are orthogonal to each other.
Compared to the \emph{3$\times$3} conv,
which captures only nine pixels around the center (shown in Fig.\ref{Fig:stripe}(a)),
the vertical and horizontal stripe convolutions have \emph{11$\times$3} and \emph{3$\times$11} receptive field,
as shown in Fig.\ref{Fig:stripe}(b).
Specifically, for a contour pixel with arbitrary orientation,
its tangent direction can be decomposed into vertical and horizontal directions.
Contexts along orthogonal directions make varied amount of contributions in expressing the orientation representation.
Thus, tendency of the extended contour and occluded relationship of bilateral scenes are recognized.

In addition, two main advantages are achieved.
First, the large receptive field aggregates contextual information of object to determine the depth order,
which is without large memory consuming.
Second, although the slope of the edge is not exactly perpendicular or parallel to the ground,
one of the stripe convolutions can successfully perceive the foreground and background objects.
After the concatenation of the two orthogonal convolution modules, we apply the \emph{3$\times$3} conv to refine the features.

\subsection{Loss Function}

{\bf{Occlusion Edge:}}
Occlusion edge characterizes depth discontinuity between regions, reflecting as the boundary between objects. Given a set of training images $\Psi = \{I_1,I_2,\dots,I_N\}$, the corresponding ground truth edge of the $k$-th input image at pixel $p$ is $E_k(p\mid I)\in \{0,1\}$ and we denote $\overline{E}_k(p\mid I,W)\in [0,1]$ as its network output, indicating the computed edge probability.

{\bf{Occlusion Orientation:}}
Occlusion orientation indicates the tangent direction of the edge using the left rule (i.e. the foreground area is on the left side of the background area). Following the mathematical definition above, for the $k$-th input image, its orientation ground truth at pixel $p$ is $O_k(p\mid I)\in (-\pi,\pi]$. The regression prediction result of orientation path is $\overline{O}_k(p\mid I,W)\in (-\pi,\pi]$.

{\bf{Occlusion Relationship:}}
During the testing phase, we first refine the $\overline{E}_k$ by conducting non-maximum suppression $\hat{E}_k = NMS(\overline{E}_k)$. The nonzero pixels of sharpened $\hat{E}_k$ form the binary matrix $M_k = sign(\hat{E}_k)$. We then perform element-wise product of $M_k$ and orientation map $O_k$, obtain refined orientation map $\hat{O}_k = M_k\circ O_k$. Finally, following [11], we adjust the $\hat{O}_k$ to the tangent direction of ${E}_k$ and gain the final occlusion edge map.

{\bf{Loss Function:}}
Following \cite{Wang2018DOOBNet}, we use the following loss function to supervise the training of our network.
\begin{equation}
\label{Eq:loss_all}
l(W)=\frac{1}{M}(\sum\limits_{j}{\sum\limits_{i}{AL({{y}_{i}},{{{e}}_{i}})+\lambda \sum\limits_{j}{\sum\limits_{i}{SL(f({\overline{a}_{i}},{{{a}}_{i}})}}}})
\end{equation}

The parameters include: collection of all standard network layer parameters ($W$), predicted edge value at pixel $i$ ($y_i\in [0,1]$), mini-batch size ($M$), image serial number in a mini-batch ($j$),the Attention Loss ($AL$), the Smooth $L_1$ Loss ($SL$) \cite{Wang2018DOOBNet}.

\section{Experiments}
\label{sec:Experiments}
In this section, abundant experiments are demonstrated to validate the performance of the proposed OFNet. 
Further, we present some ablation analyses for discussions of the network design choices.

\subsection{Implementation Details}
\label{ssec:details}

{\bf{Dataset:}}
Our method is evaluated on two challenging datasets: PIOD \cite{Peng2016DOC} and BSDS ownership \cite{Ren2006Figure}.
The PIOD dataset is composed of 9,175 training images and 925 testing images. 
Each image is annotated with ground truth object instance edge map and its corresponding orientation map.
The BSDS ownership dataset includes 100 training images and 100 testing images of natural scenes.
Following \cite{Wang2018DOOBNet}, all images in the two datasets are randomly cropped to \emph{320$\times$320} during training while retaining their original sizes during testing.

{\bf{Initialization:}}
Our network is implemented in Caffe \cite{Jia2014Caffe} and finetuned from an initial pretrained \emph{Res50} model. All convolution layers added are initialized with $``$msra$"$ \cite{He2015Delving}.

{\bf{Evaluation Criteria:}}
Following \cite{Peng2016DOC}, we compute precision and recall of the estimated occlusion edge maps (i.e.OPR) by performing three standard evaluation metrics: fixed contour threshold (ODS), best threshold of each image (OIS) and average precision (AP). Notably, the orientation recall is only calculated at the correctly detected edge pixels. Besides, the above three metrics are also used to evaluate the edge map after \emph{NMS}.

\begin{table}[t]
\small
\caption{OPR results on PIOD (left) and BSDS ownership dataset (right). \ding{172}-\ding{176} represent SRF-OCC \cite{Teo2015Fast}, DOC-HED \cite{Peng2016DOC}, DOC-DMLFOV \cite{Peng2016DOC}, DOOBNet \cite{Wang2018DOOBNet} and ours, respectively.  $\dagger$ refers to GPU running time. Red bold type indicates the best performance, blue bold type indicates the second best performance (the same below).
}
\vspace{-4mm}
\renewcommand\arraystretch{1.3}
\begin{center}
\begin{tabular}{ p{0.3cm} | p{0.4cm} p{0.4cm} p{0.4cm} p{0.9cm} | p{0.4cm} p{0.4cm} p{0.4cm} p{0.9cm}}
\hline
\emph{ }  &ODS &OIS &AP &FPS &ODS &OIS &AP &FPS\\
\hline
\ding{172}   &$.268$ &$.286$ &$.152$ &$0.018$ &$.419$ &$.448$ &$.337$ &$0.018$\\
\hline
\ding{173}   &$.460$ &$.479$ &$.405$ &$18.3\dagger$ &$.522$ &$.545$ &$.428$ &$19.6\dagger$\\
\ding{174}   &$.601$ &$.611$ &$.585$ &$18.9\dagger$ &$.463$ &$.491$ &$.369$ &$21.1\dagger$\\
\hline
\ding{175}   &$\textbf{{\color{blue}.702}}$ &$\textbf{{\color{blue}.712}}$ &$\textbf{{\color{blue}.683}}$ &$26.7\dagger$ &$\textbf{{\color{blue}.555}}$ &$\textbf{{\color{blue}.570}}$ &$\textbf{{\color{blue}.440}}$ &$25.8\dagger$\\
\ding{176}   &$\textbf{{\color{red}.718}}$ &$\textbf{{\color{red}.728}}$ &$\textbf{{\color{red}.729}}$ &$28.3\dagger$ &$\textbf{{\color{red}.583}}$ &$\textbf{{\color{red}.607}}$ &$\textbf{{\color{red}.501}}$ &$27.2\dagger$\\
\hline

\end{tabular}
\label{tab:OPR}
\end{center}
\vspace{-4mm}
\end{table}

\begin{table}[t]
\small
\caption{EPR results on PIOD (left) and BSDS ownership dataset (right). 
}
\vspace{-4mm}
\renewcommand\arraystretch{1.3}
\begin{center}
\begin{tabular}{ l | l  l  l | l  l  l}
\hline
\emph{ }  &ODS &OIS &AP &ODS &OIS &AP\\
\hline
\ding{172}   &$.345$ &$.369$ &$.207$ &$.511$ &$.544$ &$.442$\\
\hline
\ding{173}   &$.509$ &$.532$ &$.468$ &$\textbf{{\color{blue}.658}}$ &$\textbf{{\color{blue}.685}}$ &$\textbf{{\color{red}.602}}$\\
\ding{174}   &$.669$ &$.684$ &$.677$ &$.579$ &$.609$ &$.519$\\
\hline
\ding{175}   &$\textbf{{\color{blue}.736}}$ &$\textbf{{\color{blue}.746}}$ &$\textbf{{\color{blue}.723}}$ &$.647$ &$.668$ &$.539$\\
\ding{176}   &$\textbf{{\color{red}.751}}$ &$\textbf{{\color{red}.762}}$ &$\textbf{{\color{red}.773}}$ &$\textbf{{\color{red}.662}}$ &$\textbf{{\color{red}.689}}$ &$\textbf{{\color{blue}.585}}$\\
\hline

\end{tabular}
\label{tab:EPR}
\end{center}
\vspace{-4mm}
\end{table}

\subsection{Evaluation Results}
\label{ssec:eva}

{\bf{Quantitative Performance:}}
We evaluate our approach with comparisons to the state-of-the-art algorithms including SRF-OCC \cite{Teo2015Fast}, DOC-HED \cite{Peng2016DOC}, DOC-DMLFOV \cite{Peng2016DOC} and DOOBNet \cite{Wang2018DOOBNet}.

\begin{figure}[t]
\centering
\includegraphics[width=1\linewidth]{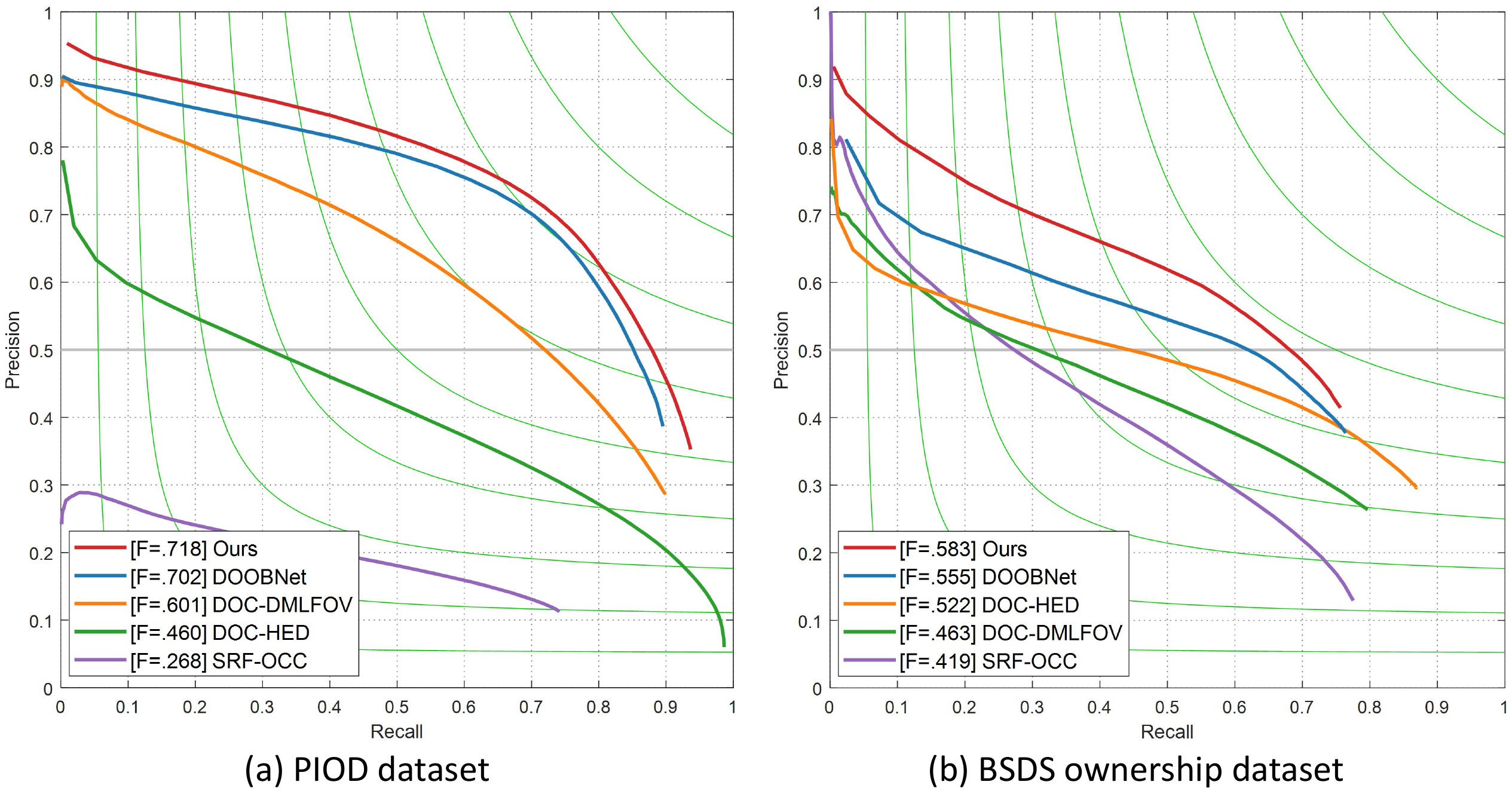}
\caption{
OPR results on two datasets.
}
\label{Fig:OPR}
\end{figure}

\begin{figure}[t]
\centering
\includegraphics[width=1\linewidth]{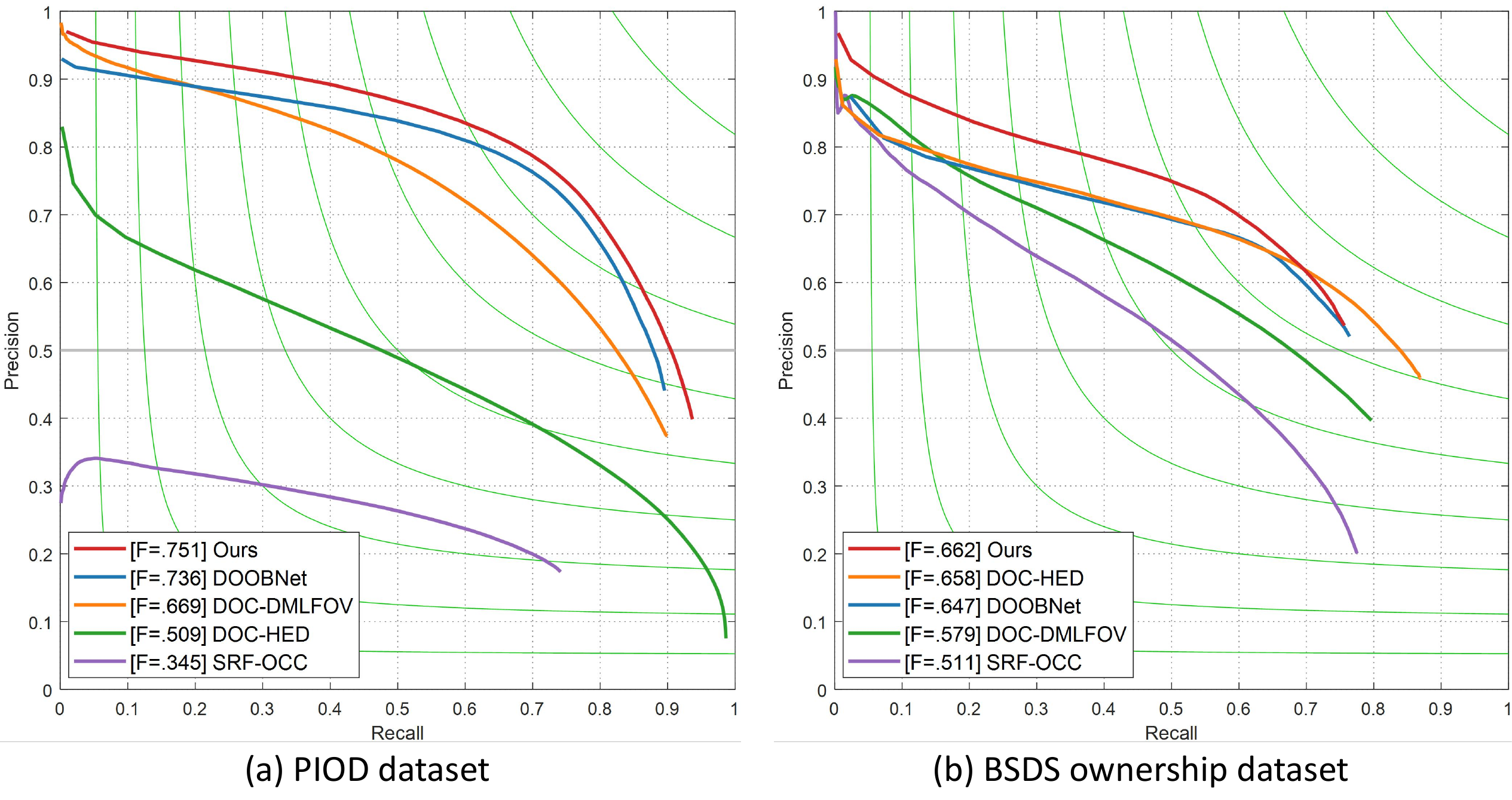}
\caption{
EPR results on two datasets.
}
\label{Fig:EPR}
\vspace{-4mm}
\end{figure}

\begin{figure*}[h]
\centering
\includegraphics[width=0.9\linewidth]{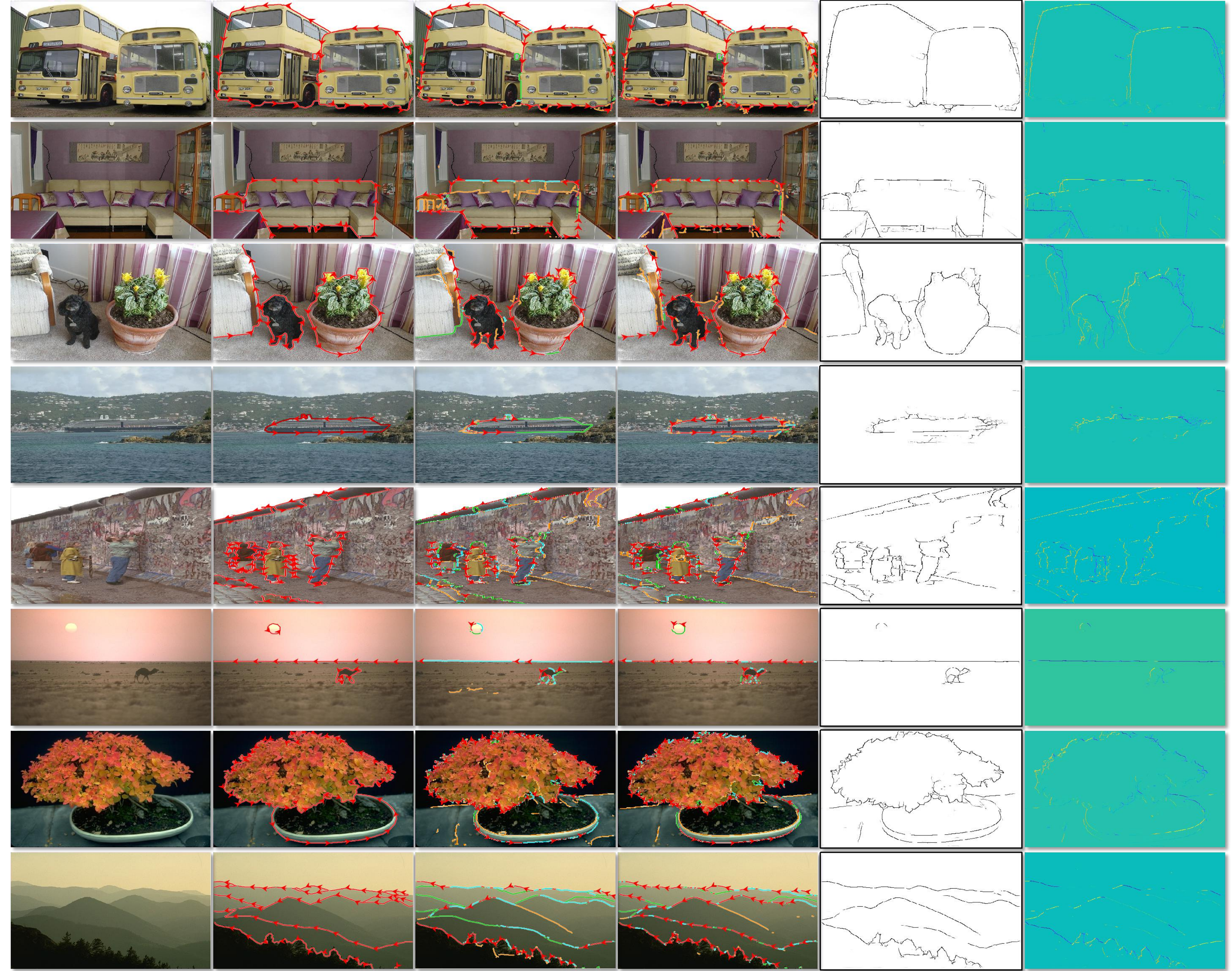}
\caption{
Example results on PIOD (first four rows) and BSDS ownership dataset (last four rows). $1^{st}$ column: input images; $2^{nd}$-$4^{th}$ columns: visualization results of ground truth, baseline and ours; $5^{th}$-$6^{th}$ columns: edge maps and orientation maps of ours.
Notably, "red" pixels with arrows: correctly labeled occlusion boundaries; "cyan": correctly labeled boundaries but mislabeled occlusion; "green": false negative boundaries; "orange": false positive boundaries (Best viewed in color). 
}
\label{Fig:quality}
\end{figure*}

As shown in Table.\ref{tab:OPR} and Fig.\ref{Fig:OPR}, our approach outperforms all other state-of-the-art methods for OPR results. Specifically, in terms of the PIOD dataset, our method performs the best,
outperforming the baseline DOOBNet of 4.6\% AP. This is due to the efficiency of extracting high semantic features of the two paths separately. Edge path succeeds in enhancing contour response and orientation path manages to perceive foreground and background relationship. Splitting these two tasks 
into two paths enables the promotion 
of the previous algorithms.
For the BSDS ownership dataset, which is difficult to train due to the small training samples,
the proposed OFNet obtains the gains of 2.8\% ODS, 3.7\% OIS and 6.1\% AP compared with the baseline DOOBNet.
Specifically, our approach increases bilateral cue between the foreground and background objects,
and fuses 
them with high semantic features to introduce clear contour, which describes the areas of the foreground and background better.
Besides, stripe convolution in our network plays an important role in harvesting the surrounding scenes of the contour. The improvement in orientation proves the effectiveness of the module.

EPR results are presented in Table.\ref{tab:EPR} and Fig.\ref{Fig:EPR}.
For the PIOD dataset, our approach superiorly performs against the other evaluated methods, surpassing DOOBNet by 5.0\% AP.
We take the distinction between edge and orientation into consideration, and extract specific features for sub-networks, respectively. For edge path,
by utilizing the contextual features, which reflect pixels constraint on the occlusion edge, our network outputs edge maps with augmented contour and less noise around. 
With location cue extracted from low-level layers, the predicted edge in our method fits 
the contour better,
thus avoiding false positive detections compared to others.
For the BSDS ownership dataset, our approach achieves the highest 
ODS as well.

{\bf{Qualitative Performance:}}
Fig.\ref{Fig:quality} shows the qualitative results on the two datasets.
The top four rows show the results of the PIOD dataset \cite{Peng2016DOC},
and the bottom four rows represent the BSDS ownership dataset \cite{Ren2006Figure}.
The first column to the sixth column show the original RGB image from datasets,
ground truth, the result predicted by DOOBNet \cite{Wang2018DOOBNet},
the result predicted by the proposed OFNet, the detected occlusion edge and the predicted orientation, respectively.
In the resulting image,
the right side of the arrow direction is the background,
and the left side corresponds to the foreground area.

In detail, the two occluded buses in the first row have similar appearances. Thus, it is hard to detect the dividing line between them, just as our baseline DOOBNet fails. However, our method detects the occlusion edge consistently.
In the second row,
the occlusion relationship between the wall and the sofa is easy to predict failure.
Instead of the small receptive field, which is difficult to perceive objects with large-area pure color, 
our method with sufficient receptive field correctly predicts the relationship.
The third scene is similar to the second row.
Compared with the baseline,
our method predicts the relationship between the sofa and the ground correctly.
In the fourth row, the color of the cruise ship is similar to the hill behind, which is not detected by the baseline.
By using the low-level edge cues, our method accurately locates the contour of the ship.
The fifth row shows people under the wall,
and the orientation cannot be correctly detected due to the low-level features in the textureless areas.
Our method correctly infers the relationship by using the high-level bilateral feature.
The last three scenes have the same problem as the third row, i.e, the object with a large region of pure color.
Our method outperforms others in this situation by a large margin, which proves the effectiveness of our designed modules.


\subsection{Ablation Analysis}
\label{ssec:ablation}

{\bf{One-branch or Multi-branch Sub-networks:}}
To evaluate that our method provides different high-level features for different sub-tasks,
an existing method \cite{Wang2018DOOBNet},
which adopts a single flow architecture by sharing high-level features,
is used to be compared with our method.
As shown in Table.\ref{tab:ablation1},
the high-level features for two paths promote the correctness of occlusion relationship.
In addition, each path is individually trained for comparison,
validating the help of
occlusion cue
for orientation prediction in our method.


\begin{table}[t]
\footnotesize
\caption{Experimental results of baseline DOOBNet \cite{Wang2018DOOBNet}, baseline with split decoder, baseline with single stream sub-network and our approach. The experiments are conducted on the PIOD dataset (the same below). 
}
\vspace{-4mm}
\renewcommand\arraystretch{1.2}
\begin{center}
\begin{tabular}{ p{3cm} | p{0.4cm} p{0.4cm} p{0.6cm} | p{0.4cm} p{0.4cm} p{0.4cm}}
\hline
\emph{Methods}  &ODS &OIS &AP &ODS &OIS &AP\\
\hline
Baseline                    &$.736$ &$.746$ &$.723$ &$.702$ &$.712$ &$.683$\\
Baseline(split decoder)     &$.720$ &$.735$ &$.694$
&$.702$ &$.712$ &$.683$\\
\hline
Single edge stream          &$.739$ &$.750$ &$.685$
&$-$ &$-$ &$-$\\
Single ori stream           &$-$ &$-$ &$-$
&$.705$ &$.716$ &$.674$\\
\hline
Ours                        &$\textbf{{\color{red}.751}}$ &$\textbf{{\color{red}.762}}$ &$\textbf{{\color{red}.773}}$
&$\textbf{{\color{red}.718}}$ &$\textbf{{\color{red}.728}}$ &$\textbf{{\color{red}.729}}$\\

\hline
\end{tabular}
\label{tab:ablation1}
\end{center}
\vspace{-4mm}
\end{table}

\begin{table}[t]
\footnotesize
\caption{Experimental results of our model without low-cues, without edge high-cues, without orientation high-cues and our model.
}
\vspace{-3mm}
\renewcommand\arraystretch{1.2}
\begin{center}
\begin{tabular}{ p{3cm} | p{0.4cm} p{0.4cm} p{0.6cm} | p{0.4cm} p{0.4cm} p{0.4cm}}
\hline
\emph{Methods}  &ODS &OIS &AP &ODS &OIS &AP\\
\hline
Ours(w/o low-cues)            &$.746$ &$.758$ &$.764$
&$.715$ &$.722$ &$.715$\\
Ours(w/o edge high-cues)      &$.742$ &$.753$ &$.758$
&$.709$ &$.717$ &$.698$\\
Ours(w/o ori high-cues)       &$.743$ &$.756$ &$.759$
&$.711$ &$.719$ &$.703$\\
\hline
Ours                          &$\textbf{{\color{red}.751}}$ &$\textbf{{\color{red}.762}}$ &$\textbf{{\color{red}.773}}$
&$\textbf{{\color{red}.718}}$ &$\textbf{{\color{red}.728}}$ &$\textbf{{\color{red}.729}}$\\

\hline
\end{tabular}
\label{tab:ablation2}
\end{center}
\vspace{-6mm}
\end{table}

{\bf{Necessity for Each Feature:}}
In order to verify the role of various low-level and high-level features,
each feature is removed to construct an independent variant for evaluation,
as shown in Table.\ref{tab:ablation2}.
Intuitively,
if the low-level features for edge path are removed,
the occlusion edge is difficult to be accurately located.
If the high-level features for edge path are removed,
the occlusion edge is failed to be detected consistently.
Furthermore,
if the high-level features for orientation path are removed,
although the occlusion edge could be detected accurately and consistently,
the ability to reason occlusion relationship reduces sharply.
The intrinsic reason is that the \textbf{MCL} perceives the bilateral cue around the contour,
and affirms the foreground and background relationship.
The bilateral feature plays an important role in occlusion relationship reasoning.



{\bf{Proportion of Bilateral and Contour Features:}}
The bilateral feature provides relative depth edgewise,
and occlusion cue supplies the location of the boundary.
We fuse them with various channel ratios to best refine the range of the foreground and background.
The proportion of bilateral and
occlusion
features determines the effectiveness of the fusion.
Table.\ref{tab:ablation3} reveals various experimental results with different proportions of two features.
Experiments prove that fusing bilateral feature and
occlusion
feature with \emph{64:16} channel ratio in the \textbf{BRF} outperforms others.
It reveals that bilateral feature plays a more important role in the fusion operation.
Occlusion cue
mainly plays an auxiliary role, which distinguishes the region of foreground and background.
However, when the bilateral feature occupies an excess proportion,
the boundary will be ambiguous,
blurring the boundary between foreground and background,
which causes a negative impact on the effect.

\begin{table}[t]
\footnotesize
\caption{Experimental results of bilateral feature and 
occlusion
feature with various fusion ratio in \textbf{BRF} module.
}
\renewcommand\arraystretch{1.2}
\begin{center}
\begin{tabular}{ p{2cm} | p{0.4cm} p{0.4cm} p{0.6cm} | p{0.4cm} p{0.4cm} p{0.4cm}}
\hline
\emph{Scale}  &ODS &OIS &AP &ODS &OIS &AP\\
\hline
scale = \emph{16:16}               &$.742$ &$.752$ &$.749$ 
&$.710$ &$.719$ &$.703$\\
scale = \emph{32:16}               &$.741$ &$.754$ &$.759$
&$.712$ &$.722$ &$.709$\\
scale = \emph{48:16}               &$.744$ &$.758$ &$.765$
&$.715$ &$.726$ &$.717$\\
\hline
scale = \emph{64:16}               &$\textbf{{\color{red}.751}}$ &$\textbf{{\color{red}.762}}$ &$\textbf{{\color{red}.773}}$
&$\textbf{{\color{red}.718}}$ &$\textbf{{\color{red}.728}}$ &$\textbf{{\color{red}.729}}$\\
\hline
scale = \emph{80:16}               &$.747$ &$.757$ &$.764$
&$.715$ &$.726$ &$.722$\\

\hline
\end{tabular}
\label{tab:ablation3}
\end{center}
\vspace{-4mm}
\end{table}

\begin{table}[t]
\footnotesize
\caption{Experimental results of stripe convolutions with different aspect ratios.
}
\renewcommand\arraystretch{1.2}
\begin{center}
\begin{tabular}{ p{2cm} | p{0.4cm} p{0.4cm} p{0.6cm} | p{0.4cm} p{0.4cm} p{0.4cm}}
\hline
\emph{Scale}  &ODS &OIS &AP &ODS &OIS &AP\\
\hline
conv = \emph{3$\times$3}               &$.746$ &$.753$ &$.754$ &$.712$ &$.719$ &$.694$\\
conv = \emph{3$\times$5}               &$.747$ &$.755$ &$.760$
&$.712$ &$.720$ &$.696$\\
conv = \emph{3$\times$7}               &$.747$ &$.754$ &$.758$
&$.713$ &$.721$ &$.699$\\
conv = \emph{3$\times$9}               &$.750$ &$.759$ &$.767$
&$.716$ &$.723$ &$.712$\\
\hline
conv = \emph{3$\times$11}             &$\textbf{{\color{red}.751}}$ &$\textbf{{\color{red}.762}}$ &$\textbf{{\color{red}.773}}$
&$\textbf{{\color{red}.718}}$ &$\textbf{{\color{red}.728}}$ &$\textbf{{\color{red}.729}}$\\

\hline
\end{tabular}
\label{tab:ablation4}
\end{center}
\vspace{-6mm}
\end{table}

{\bf{Plain or Stripe Convolution:}}
To evaluate the effect of stripe convolution for occlusion relationship reasoning,
the stripe-based convolution variants with different aspect ratios are employed to make comparisons.
As shown in Table.\ref{tab:ablation4},
intuitively,
even if the slope of the edge is not in a horizontal or vertical direction,
the convolution kernels possess large receptive field and tend to learn the cues of both directions, respectively.
Nevertheless, the larger convolution layer takes up too much computation cost,
which increases the number of parameters.
Consequently, the stripe convolutions in orthogonal directions extract the tendency of edges and bilateral cue around the contour.

\section{Conclusion}
In this paper, we present a novel OFNet, which shares the
occlusion cue
from the decoder and separately acquires the contextual features for specific tasks.
Our algorithm builds on top of the encoder-decoder structure and side-output utilization. For learning the bilateral feature, an \textbf{MSL} is proposed.
Besides, a \textbf{BRF}
module is designed to apply the
occlusion cue
to precisely locate the object regions.
In addition, we utilize a stripe convolution to further aggregate features from surrounding scenes of the contour.
Significant improvement of the state-of-the-art through numerous experiments on PIOD and BSDS ownership dataset demonstrates the effectiveness of our network.

~

\noindent{\bf{Acknowledgement.}}
This work was supported by the National Natural Science Foundation of China Nos.~61703049, 61876022, 61872047, 
and the Fundamental Research Funds for the Central Universities No.~2019kfyRCPY001.

{\small
\balance
\bibliographystyle{ieee_fullname}
\bibliography{ICCV2019_arxiv}
}

\renewcommand\thesection{\Alph{section}} 

\setcounter{section}{0}

\section{Appendix}

\begin{figure*}[t]
\centering
\includegraphics[width=0.85\linewidth]{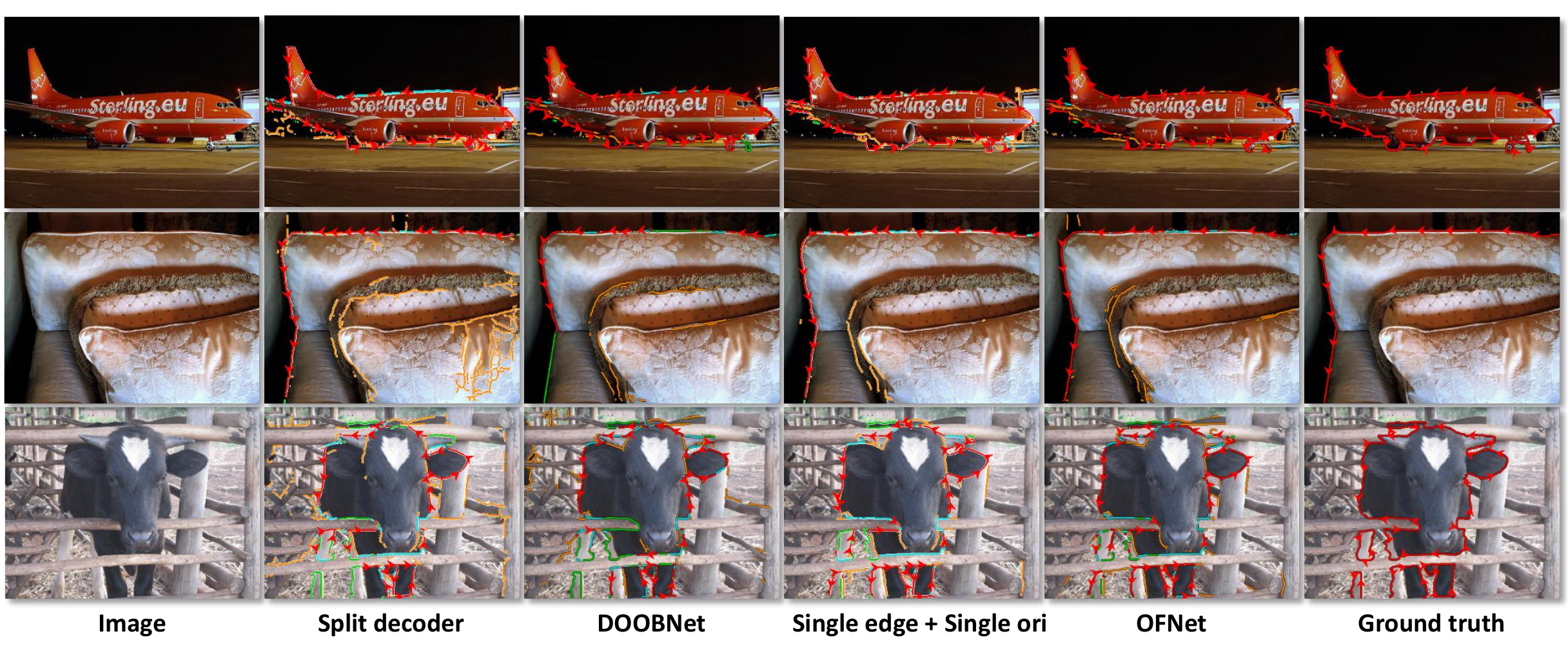}
\caption{
Occlusion relationship of various approaches.
The occlusion relationship (the red arrows) is represented by orientation $\theta\in(-\pi,\pi]$ (tangent direction of the edge), using the "left" rule where the left side of the arrow means foreground area.
Notably, "red" pixels with arrows: correctly labeled occlusion boundaries; "cyan": correctly labeled boundaries but mislabeled occlusion; "green": false negative boundaries; "orange": false positive boundaries (Best viewed in color).
Column $1^{st}$: Input image. Column $2^{nd}-5^{th}$: Output of split decoder, DOOBNet, Single edge $+$ single ori and OFNet. Column $6^{th}$: Ground truth.
}
\label{Fig:ablation1}
\end{figure*}

In this appendix material, we provide full qualitative analysis for the ablation study. The experiments are conducted on the PIOD dataset.

\subsection{One-branch or Multi-branch Sub-networks}

Previous approach DOOBNet \cite{Wang2018DOOBNet} adopts a single flow architecture by sharing decoder features that represent high-level features.
The shared decoder features reflect the contour cues,
which are necessary for both edge and orientation estimations.
Besides, edge detection and orientation detection are different in the choice of feature extraction, especially in the case of high semantic layers.
We innovatively split the features produced by side-outputs and share decoder features to fit both tasks, respectively. Fig.\ref{Fig:ablation1} reveals the effectiveness of our design.

\subsection{Necessity for Each Feature}

\begin{figure*}[t]
\centering
\includegraphics[width=0.85\linewidth]{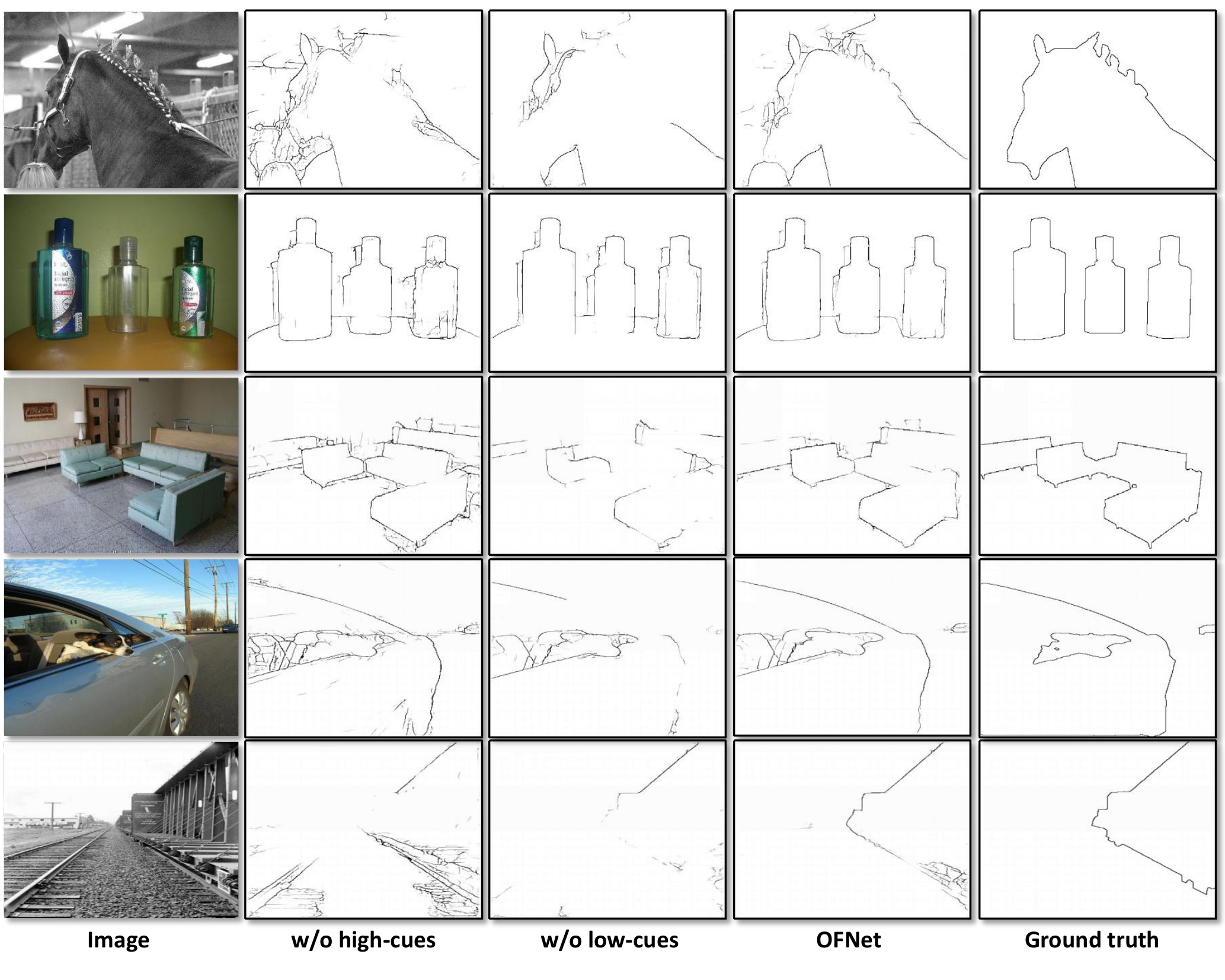}
\caption{
Edge maps of various approaches.
Column $1^{st}$: Input image. Column $2^{nd}-4^{th}$: OFNet without high-cues, OFNet without low-cues and OFNet. Column $5^{th}$: Ground truth.
}
\label{Fig:ablation2}
\end{figure*}

To verify the role of various low-level and high-level features,
each feature is removed to construct an independent variant for evaluation.
If the low-level features for edge path are removed,
the occlusion edge is difficult to be accurately located,
leading to decrease in the accuracy of occlusion relationship reasoning (shown in Fig.\ref{Fig:ablation2}(\emph{w/o low-cues})).
If the high-level feature for edge path is removed,
the occlusion edge is failed to be detected consistently,
which decreases the accuracy at a large margin (shown in Fig.\ref{Fig:ablation2}(\emph{w/o high-cues})).
By capturing spatial and contextual cues from the side-outputs respectively,
the network is able to explore specific features for individual predictions.

\subsection{Proportion of Bilateral-Contour Features}

\begin{figure*}[t]
\centering
\includegraphics[width=1\linewidth]{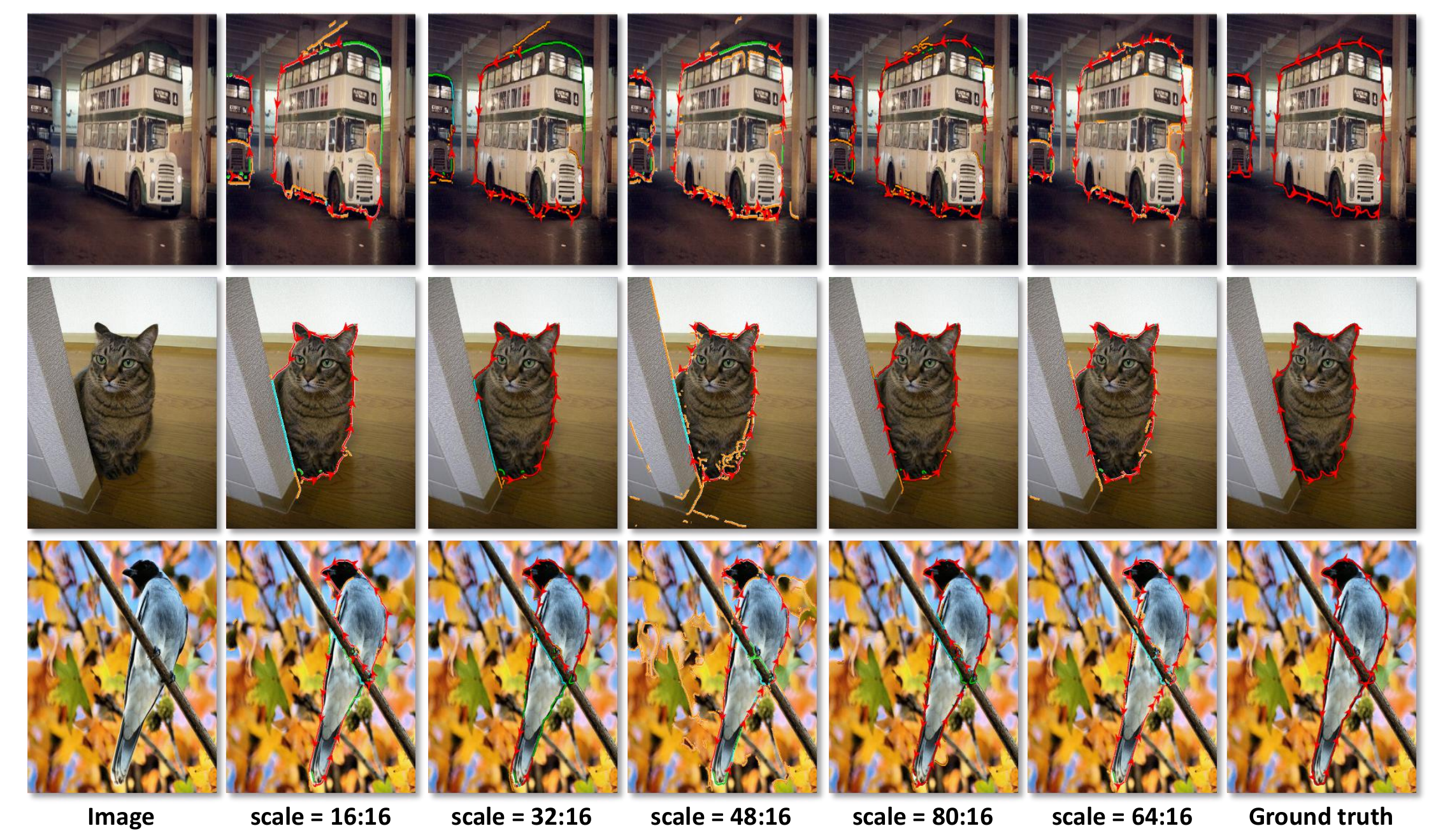}
\caption{
Occlusion relationship of various approaches.
Column $1^{st}$: Input image. Column $2^{nd}-6^{th}$: Fusing bilateral feature and occlusion feature with 16:16, 32:16, 48:16, 80:16 and 64:16 channel ratio, respectively. Column $7^{th}$: Ground truth.
}
\label{Fig:ablation3}
\end{figure*}

Previous works utilize inappropriate feature maps to predict the orientation, which reflects as the characteristic of the edge outline. The features of orientation on both sides of the contour are filtered gradually which are adversely affected by the edge prediction. We take advantage of an \textbf{MCL} to perceive the bilateral cues around the contours, and affirm the foreground and background relationship. 
As shown in Fig.\ref{Fig:ablation3}, fusing bilateral feature and
occlusion feature with \emph{64:16} channel ratio in the \textbf{BRF} outperforms others.

\subsection{Plain or Stripe Convolution}

Plain convolutions perceive information about surrounding small areas.
To extract the tendency of edges to extend and bilateral cues around contours,
we employ stripe convolutions in orthogonal directions.
The convolution kernels possess large receptive field and tend to learn the cues of both directions, respectively.
We test stripe convolution kernels with different aspect ratios,
which are exhibited in Fig.\ref{Fig:ablation3}.
The larger convolution layer takes up too much computation cost,
which increases the number of parameters.
We evaluate the performance of the model with \emph{11$\times$11 conv} on PIOD and BSDS datasets, 
the \emph{EPR} (left) and \emph{OPR} (right) are reported in Table.\ref{tab:abl}. 
Compared with \emph{3$\times$11 conv}, the model with \emph{11$\times$11 conv} achieves limited improvement, 
while it increases about \emph{50$\%$} gpu memory usage (\emph{10031MB} to \emph{14931MB}).

\begin{figure*}[t]
\centering
\includegraphics[width=1\linewidth]{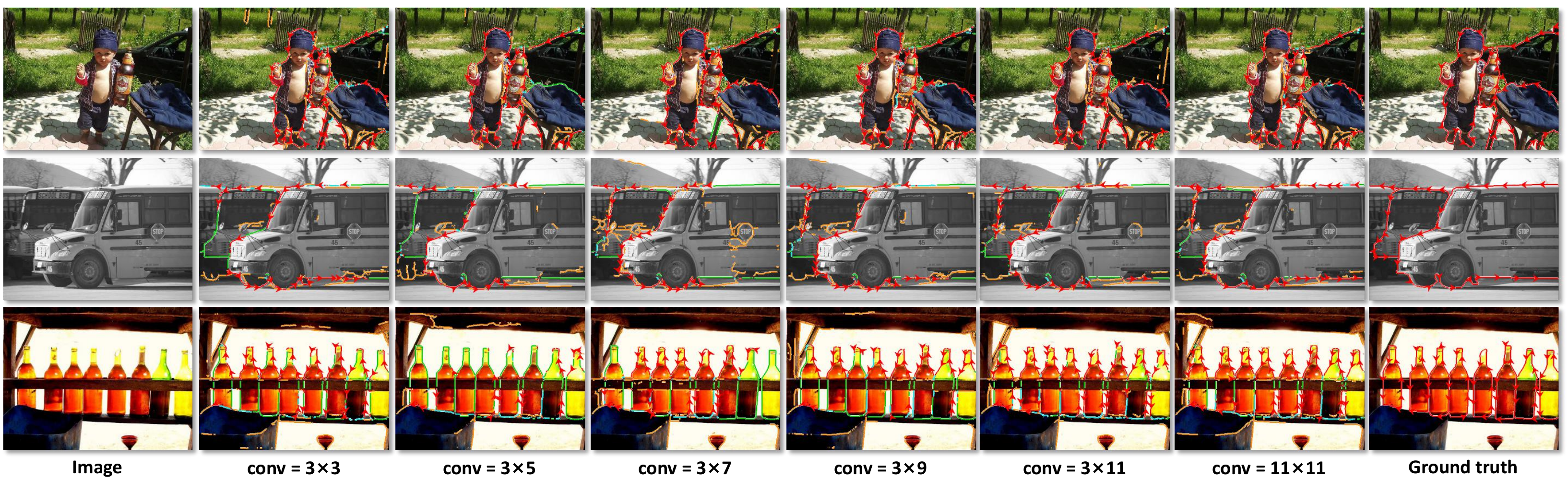}
\caption{
Edge maps of various approaches.
Column $1^{st}$: Input image. Column $2^{nd}-7^{th}$: conv kernel size = 3$\times$3, 3$\times$5, 3$\times$7, 3$\times$9, 3$\times$11 and 11$\times$11. Column $8^{th}$: Ground truth.
}
\label{Fig:ablation4}
\end{figure*}

\begin{table}[h]
\scriptsize
\caption{Results of our model with different conv kernel sizes.}
\vspace{-6mm}
\renewcommand\arraystretch{1.2}
\begin{center}
\begin{tabular}{ c | c | c  c c | c c c}
\hline
\textbf{Dataset} &\textbf{Scale}  &\textbf{ODS} &\textbf{OIS} &\textbf{AP} &\textbf{ODS} &\textbf{OIS} &\textbf{AP}\\
\hline
\multirow{2}*{\emph{PIOD}}
&\emph{conv = 3$\times$11}               &$.751$ &$.762$ &$.773$
&$.718$ &$.728$ &$.729$\\
&\emph{conv = 11$\times$11}           &$.753$ &$.764$ &$.776$
&$.719$ &$.730$ &$.732$\\
\hline
\multirow{2}*{\emph{BSDS}}
&\emph{conv = 3$\times$11}               &$.662$ &$.689$ &$.585$
&$.583$ &$.607$ &$.501$\\
&\emph{conv = 11$\times$11}           
&$.663$ &$.690$ &$.587$
&$.585$ &$.608$ &$.503$\\

\hline
\end{tabular}
\label{tab:abl}
\end{center}
\vspace{-8.8mm}
\end{table}

\end{document}